\def\year{2020}\relax
\documentclass[letterpaper, onecolumn]{article} 
\usepackage{aaai20}  
\usepackage{times}  
\usepackage{helvet} 
\usepackage{courier}  
\usepackage[hyphens]{url}  
\usepackage{threeparttable}
\usepackage{graphicx} 
\usepackage{graphicx}  
\usepackage{amsmath}
\usepackage{amssymb}
\title{Expressing Objects just like Words: Recurrent Visual Embedding for\\ Image-Text Matching}
\author{ \Large \textbf{Tianlang Chen, Jiebo Luo}\\ 
Department of Computer Science\\
University of Rochester\\ 
tchen45@cs.rochester.edu, jluo@cs.rochester.edu\\
}
\begin{document}

\maketitle

\begin{abstract}
Existing image-text matching approaches typically infer the similarity of an image-text pair by capturing and aggregating the affinities between the text and each independent object of the image. However, they ignore the connections between the objects that are semantically related. These objects may collectively determine whether the image corresponds to a text or not. To address this problem, we propose a Dual Path Recurrent Neural Network (DP-RNN) which processes images and sentences symmetrically by recurrent neural networks (RNN). In particular, given an input image-text pair, our model reorders the image objects based on the positions of their most related words in the text. In the same way as extracting the hidden features from word embeddings, the model leverages RNN to extract high-level object features from the reordered object inputs. We validate that the high-level object features contain useful joint information of semantically related objects, which benefit the retrieval task. To compute the image-text similarity, we incorporate a Multi-attention Cross Matching Model into DP-RNN. It aggregates the affinity between objects and words with cross-modality guided attention and self-attention. Our model achieves the state-of-the-art performance on Flickr30K dataset and competitive performance on MS-COCO dataset. Extensive experiments demonstrate the effectiveness of our model.
\end{abstract}

\section{Introduction}

Image-text matching involves the task to measure the similarity between an image and a text. By image-text matching, a system can retrieve the top corresponding images of a sentence query, or retrieve the top corresponding sentences of an image query.

Currently, approaches based on bottom-up attention achieve the best performance on this task. Given an image and a text as the input, these approaches first figure out the affinity between image objects and words. After getting these local matching informative snippets, they predict the image-text similarity by aggregating them appropriately. Approaches based on bottom-up attention perform much better than directly extracting the global image and text features and compute their similarity. 

\begin{figure}[!t]
\centering
\includegraphics[width=0.69\columnwidth]{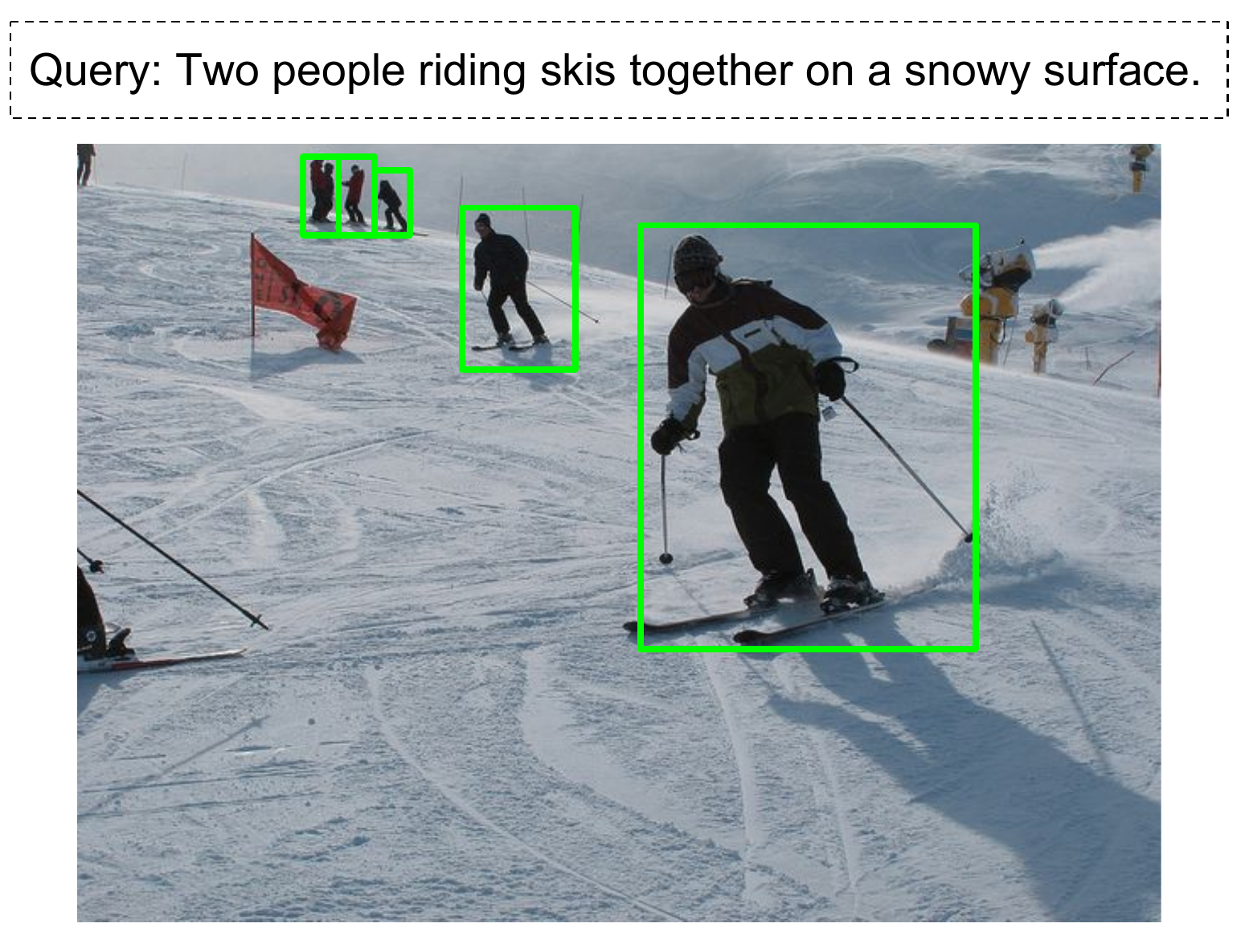}
\caption{An example to illustrate that the mismatch of an image-text pair cannot be detected by any single object.}
\label{fig:problem}
\end{figure}

For the text branch, high-level semantic features are extracted from RNN, which captures the joint information of adjacent and semantically related words. This becomes a bridge to connect the ``local'' words and the ``global'' text. However, for the image branch, current approaches based on bottom-up attention consider each image object independently. Typically, they compute the image-text similarity as the average or weighted average of the similarities between the text and each image object. In this process, essential joint information of semantically related objects cannot be extracted, preventing the model from making more accurate predictions. Given the example of  Figure~\ref{fig:problem}, each object in the green box will get a high predicted similarity score to the query because it represents an important element of the query, i.e. ``people''. However, only by jointly modeling these objects can the model predict that the image does not correspond to the text, because the objects do not match the description of ``two people ride skis together''.

To address this problem, we propose a novel Dual Path Recurrent Neural Network (DP-RNN) for image-text matching. In particular, we consider the relation between image and objects in accordance with the relation between sentence and words. Given an image-text pair as the input, DP-RNN ``reads'' the image objects in the corresponding order as the text indicates. More concretely, for each object, the model first searches its most related word in the text, regardless of whether the image and text match. After that, the model reorders these objects based on the positions of their most matching words in the text. Finally, the reordered object features are fed into RNN, from which the high-level object embeddings are extracted to capture the joint information of semantically related objects. We refer to the above process as recurrent visual embedding. Given a training batch containing $s$ images and texts, our model needs to perform recurrent visual embedding for $s^{2}$ image-text pairs, thus suffering from extremely high computational complexity. Therefore, we design an effective pair early-selection strategy to adaptively filter out insignificant image-text pairs.   

After extracting the high-level word and object features, we propose a Multi-attention Cross Matching Model to predict the image-text similarity. More concretely, following the idea of \cite{lee2018stacked}, we leverage a cross attention model to symmetrically predict the similarity of each word to the image (i.e. image-word similarity) and the similarity of each object to the text (i.e. object-text similarity). In this process,  the model attends differentially to image objects and words using both as the context to each other. This can be considered as cross-modality guided attention. Furthermore, we feed two self-attention modules to predict the weight of each word and each object. On one hand, the model figures out the word-oriented image-text similarity by the predicted word weights and the image-word similarity. On the other hand, the model figures out the object-oriented image-text similarity by the predicted object weights and the object-text similarity. The final image-text similarity is the average of these two types of similarities.

Our contributions are summarized as follows: 
\begin{itemize}
    \item  We design and apply recurrent neural networks with a pair early-selection strategy for visual embeddings. It adaptively extracts effective joint information from semantically related objects by object reordering.
    \item We propose a Multi-attention Cross Matching Model to compute similarity to further improve the image-text matching performance.
\end{itemize}

\section{Related Work}

Image-text matching has received much attention in recent years due to the advances in computer vision and natural language processing. Early image-text matching approaches capture visual-textual correspondence at the level of
image and text. Frome et al. \cite{frome2013devise} propose the first visual-semantic embedding framework for image-text matching. The image feature and text feature are extracted by CNN and Skip-Gram Language Model \cite{mikolov2013efficient}, respectively. Ranking loss is then implemented for similarity learning. Kiros et al. \cite{kiros2014unifying}, on the other hand, encode text by RNN and design a hinge-based triplet ranking loss to train the model. Faghri et al. \cite{faghri2017vse++} leverage hard negatives in the triplet loss to train the model, it shows better performance than randomly sampling negative image-text pairs. Gu et al. \cite{gu2018look} integrate a generative module to generate the corresponding images from the text feature and guide the model to learn a better representation space.

Recent successes of attention models for visual-textual learning tasks, such as for image captioning \cite{xu2015show,lu2017knowing,you2016image,pedersoli2017areas} and visual question answering (VQA) \cite{yu2017multi,lu2016hierarchical,yang2016stacked,kim2016multimodal}, motivate researchers to solve image-text matching at the level of image regions and words. Huang et al. \cite{huang2017instance} incorporate a multimodal context-modulated attention scheme that can selectively attend to a pair of instances of image and sentence at each time step. Li et al. \cite{li2017identity} design a latent co-attention mechanism to relate each word to the corresponding image regions. Similarly, Nam et al. \cite{nam2017dual} propose Dual Attention Networks that attend to both specific regions in images and words in text through multiple steps. Because of the restriction of CNN, each image is typically divided into a fixed number of regions (e.g $7 \times 7$) of the same shape and size. This prevents models from  matching between words and small image objects more accurately.

Anderson et al. \cite{anderson2018bottom} propose bottom-up attention for the task of image captioning
and VQA. It directly constructs the connection between words and image objects extracted by the object detection model \cite{ren2015faster}. In accordance with image captioning
and VQA, approaches based on bottom-up attention remarkably improve the performance of image-text matching. Niu et al. \cite{niu2017hierarchical} propose a Hierarchical LSTM model to exploit the hierarchical relations between objects and image, as well as words and text. Huang et al. \cite{huang2018learning} predict an image's semantic concepts, including objects, properties and actions to construct more accurate connection between visual and textual modality. Lee et al. \cite{lee2018stacked} propose a stacked cross attention model to discover all possible alignments between image objects and text words, and predict image-text matching with two complimentary formulations. Following their work, we propose DP-RNN which incorporates recurrent neural networks to extract visual features. It should be noticed that \cite{niu2017hierarchical} also embeds objects by a recurrent network structure. However, for each phrase of the text, they only find and embed one object with this phrase, which limits the model's capacity to extract the joint information of all possible semantically related objects. In addition, compared with \cite{huang2018learning} that also considers the order of image concept, we assign a dynamic object order for an image based on the input text and  achieve better performance.

\section{Method}
\begin{figure*}[!t]
\centering
\includegraphics[width=1.95\columnwidth]{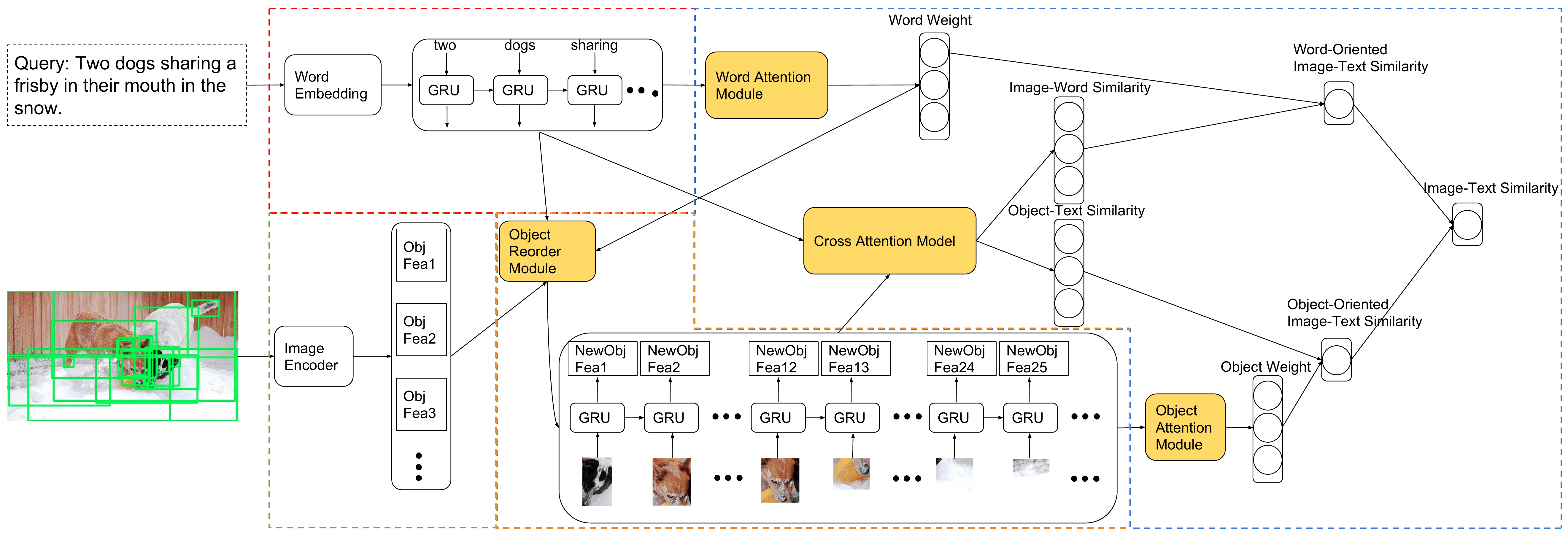}
\caption{The architecture of the proposed DP-RNN. It contains an image encoder (green), a text encoder (red), a recurrent visual embedding module (orange) and a Multi-attention Cross Matching Model (blue) to predict the image-text similarity. The objects are reordered based on the positions of their most related words in the text.}
\label{fig:model}
\end{figure*}

In this section, we formally present our Dual Path Recurrent Neural Network (DP-RNN) model. Specifically, given an image-text pair as input, the model aims to predict the pair's similarity by mapping the image objects and words into a common embedding space. The overall architecture of the proposed model is shown in Figure~\ref{fig:model}. We first introduce the overall design of DP-RNN without recurrent visual embedding, including the model's architecture and training loss function. After that, we present recurrent visual embedding as a core module of DP-RNN for better object feature extraction. In the end, we present the training strategy of DP-RNN and propose an effective pair early-selection strategy to filter out insignificant image-text pairs, which sharply reduces the computation complexity of recurrent visual embedding.

\subsection{Cross Matching with Multi-attention} \label{sec:basic}

Given an input image $I$, we pay attention to the level of image object. Following Anderson et al. \cite{anderson2018bottom}, our image encoder utilizes Faster R-CNN \cite{ren2015faster} to propose salient regions, each with an associated feature vector from ResNet-101 \cite{he2016deep}. Considering each salient region as an object, for image $I$, we extract a set of objects $\{O_{1},O_{2},...,O_{k}\}$ with the corresponding object features $\{f_{1},f_{2},...,f_{k}\}$ where $k$ = 36. The image encoder also contains a fully-connected layer, which transforms each $f_{i} \in \{f_{1},f_{2},...,f_{k}\}$ to a $h$-dimensional vector $fo_{i}$. Different from \cite{anderson2018bottom,lee2018stacked}, we additionally feed the object position information into the model. Specifically, we extract a 4-dimensional feature that contain the normalized width, height and the central coordinate of each object $O_{i}$, we map it to a $h$-dimensional vector $po_{i}$ by a fully-connected layer with sigmoid unit. The final object feature $o_{i}$ of object $O_{i}$ is the multiplication of $fo_{i}$ and $po_{i}$. On the other hand, for the input text $T$ with word set $\{W_{1},W_{2},...,W_{n}\}$, we feed it into a text encoder. The text encoder is a combination of word-embedding module and bi-directional GRU. For each word $W_{i}$ of text $T$, it is first embedded as a $q$-dimensional word representation $x_{j}$. After that, a forward GRU of bi-directional GRU reads $T$ from $x_{1}$ to $x_{n}$:

\begin{equation}
\begin{aligned}
\overrightarrow{h_{j}} = \overrightarrow{GRU}(x_{j},h_{j-1}), j \in [1,n]
\end{aligned}
\end{equation}
and a backward GRU reads $T$ from $x_{n}$ to $x_{1}$:

\begin{equation}
\begin{aligned}
\overleftarrow{h_{j}} = \overleftarrow{GRU}(x_{j},h_{j+1}), j \in [1,n]
\end{aligned}
\end{equation}

The final word feature $w_{j}$ is defined as:

\begin{equation}
\begin{aligned}
w_{j} = \frac{\overrightarrow{h_{j}}+\overleftarrow{h_{j}}}{2}, j \in [1,n]
\end{aligned}
\end{equation}
where $w_{j}$ has the same dimension (i.e. $h$) as the object feature. 

Our model predicts the similarity of pair $(I, T)$ in a symmetrical manner. As shown in Figure~\ref{fig:model}, we feed the set of object and word features into a cross attention model, which outputs the similarity between text $T$ and each object $O_{i} \in \{O_{1},O_{2},...,O_{k}\}$, as well as the similarity between image $I$ and each word $W_{j} \in \{W_{1},W_{2},...,W_{n}\}$. We denote them as $S(i,T)$ and $S(I,j)$, respectively.

To compute $S(i,T)$, following \cite{lee2018stacked}, the affinity of object $O_{i}$ to word $W_{j}$ (denoted as $A_{i,j}$)
is defined as:

\begin{equation}
\begin{aligned}
A_{i,j}  = \frac{cossim_{+}(o_{i},w_{j})}{\sqrt{\sum_{i=1}^k (cossim_{+}(o_{i},w_{j}))^{2}}},\\
cossim_{+}(o_{i},w_{j})  = \max(\frac{o_{i}^{T}w_{j}}{\left\lVert{o_{i}}\right\rVert\left\lVert{w_{j}}\right\rVert},0)
\end{aligned}
\end{equation}

In other words, the affinity of $O_{i}$ to $W_{j}$ is high if the $cosine$ similarity between $w_{j}$ and $o_{i}$ is high while the $cosine$ similarities between $w_{j}$ and other object features are low. After that, the weighted
text feature $t_{i}$ based on $o_{i}$ is defined as:

\begin{equation}\label{equ:att}
\begin{aligned}
t_{i} & = \sum_{j=1}^n \alpha_{i,j}w_{j}, &
\alpha_{i,j} & = \frac{\exp(\lambda_{1}A_{i,j})}{\sum_{j=1}^{n}\exp(\lambda_{1}A_{i,j})}
\end{aligned}
\end{equation}

Equation~\ref{equ:att} can be considered as a cross-modality guided attention mechanism, $\alpha_{i,j}$ is the cross attention weight for $O_{i}$ to $W_{j}$. Based on the object feature $o_{i}$, the model is guided to generate the text feature $t_{i}$ that focuses more on the words with high $A_{i,j}$. $\lambda_{1}$ is the inversed temperature of the softmax function. When $\lambda_{1}$ is large, the weighted text feature almost only considers the word $W_{j}$ that holds the highest $A_{i,j}$ with object $O_{i}$.

In the end, $S(i,T)$ is defined as the $cosine$ similarity between $o_{i}$ and $t_{i}$:

\begin{equation}
\begin{aligned}
S(i,T) & = \frac{o_{i}^{T}t_{i}}{\left\lVert{o_{i}}\right\rVert\left\lVert{t_{i}}\right\rVert}
\end{aligned}
\end{equation}

In a dual form, we define the affinity of word $w_{j}$ to the object $o_{i}$ as:

\begin{equation}
\begin{aligned}
\widetilde{A}_{j,i}  = \frac{cossim_{+}(w_{j},o_{i})}{\sqrt{\sum_{j=1}^n (cossim_{+}(w_{j},o_{i}))^{2}}}
\end{aligned}
\end{equation}
and the weighted image feature $m_{j}$ based on $w_{j}$ as:

\begin{equation}
\begin{aligned}
m_{j} & = \sum_{i=1}^k \widetilde{\alpha}_{j,i}o_{i}, &
\widetilde{\alpha}_{j,i} & = \frac{\exp(\lambda_{2}\widetilde{A}_{j,i})}{\sum_{i=1}^{k}\exp(\lambda_{2}\widetilde{A}_{j,i})}
\end{aligned}
\end{equation}
and $S(I,j)$ is defined as:

\begin{equation}
\begin{aligned}
S(I,j) & = \frac{w_{j}^{T}m_{j}}{\left\lVert{w_{j}}\right\rVert\left\lVert{m_{j}}\right\rVert}
\end{aligned}
\end{equation}

On the other hand, different objects/words have different importance in an image/text to express the characteristics of the image/text. Therefore, we incorporate a word attention module and an object attention module to compute the self-attention weights of words and objects. In particular, the word attention module contains a learnable $1 \times h$  matrix $W_{w}$, and the word attention weight $a_{j}^{w}$ of word $W_{j}$ is defined as:

\begin{equation}
\begin{aligned}
a_{j}^{w} & = \frac{\exp(\beta_{w} W_{w}w_{j})}{\sum_{j=1}^{n}\exp(\beta_{w} W_{w}w_{j})}
\end{aligned}
\end{equation}
$\beta_{w}$ is also the inversed temperature to adjust the self-attention. Likewise, the image attention module contains a learnable $1 \times h$ matrix $W_{o}$, and the object attention weight $\widetilde{a}_{i}^{o}$ of object $O_{i}$ is defined as:

\begin{equation}
\begin{aligned}
\widetilde{a}_{i}^{o} & = \frac{\exp(\beta_{o} W_{o}o_{i})}{\sum_{i=1}^{k}\exp(\beta_{o} W_{o}o_{i})}
\end{aligned}
\end{equation}

In the end, the word-oriented and object-oriented image-text similarity of $(I, T)$ are defined as:
\begin{equation}
\begin{aligned}
S_{w}(I,T) & = \sum_{j=1}^{n}a_{j}^{w}S(I,j), & 
S_{o}(I,T) & = \sum_{i=1}^{k}\widetilde{a}_{i}^{o}S(i,T)
\end{aligned}
\end{equation}



Following \cite{faghri2017vse++}, we adopt the triplet ranking loss with hardest negatives in a mini-batch to train the model. For a corresponding pair $(I, T)$, the hardest negatives in a mini-batch are given by $\hat{I} = argmax_{b \neq I}S(b,T)$ and $\hat{T} = argmax_{c \neq T}S(I,c)$. The triplet loss is defined as:


\begin{equation} \label{equ:loss}
\begin{aligned}
l_{hard}(I,T) = [\gamma - S(I,T)+S(I,\hat{T})]_{+} \\ + [\gamma - S(I,T)+S(\hat{I},T)]_{+} 
\end{aligned}
\end{equation}

$\gamma$ is the margin of the triplet loss, $[x]_{+} \equiv max(x, 0)$. 
 
Following \cite{lee2018stacked}, for actual implementation, we train two separate models by applying the triplet loss on the word-oriented image-text similarity (i.e. $S(I,T) = S_{w}(I,T)$) and object-oriented image-text similarity (i.e. $S(I,T) = S_{o}(I,T)$), respectively. During the inference process, we average the predicted similarity scores of the two models as the final image-text similarity.

\subsection{Recurrent Visual Embedding} \label{sec:rvt}
We formally present the recurrent visual embedding as an effective approach to extracting the joint information of semantically related image objects. Given an input image-text pair $(I, T)$ with the same denotation, we define $P(i,j)$ as the coefficient that $W_{j}$ relates to $O_{i}$:

\begin{equation}
\begin{aligned}
P(i,j) = a_{j}^{w}(o_{i}^{T}w_{j})
\end{aligned}
\end{equation}
where $a_{j}^{w}$ is the attention weight of word $W_{j}$ defined above. For object $O_{i}$, we select the $p$th word as its most related word where $p = argmax_{j}{P(i,j)}$. After that, we reorder the objects based on the position of its most related word. In essence, the model aims to make two objects close to each other if they are semantically related in the input text's environment. Therefore, given different texts as input, the object order of a same image may change. As shown in Figure~\ref{fig:model}, given an input text, the new order of the objects will be in accordance with the order of the words in the text based on their semantic meanings.

After reordering the objects of an image, we feed the object features into another bi-directional GRU layer with the same way as the word features. Specifically, we re-denote the set of the ordered object features as $\{o_{1}^{old},o_{2}^{old},...,o_{n}^{old}\}$ so that the new object features are computed as:

\begin{equation}
\begin{aligned}
\overrightarrow{h_{i}^{new}} & = \overrightarrow{GRU}(o_{i}^{old},h_{i-1}^{new}), \\
\overleftarrow{h_{i}^{new}} & = \overleftarrow{GRU}(o_{i}^{old},h_{i+1}^{new}), \\
o_{i}^{new} & = \frac{\overrightarrow{h_{i}^{new}}+\overleftarrow{h_{i}^{new}}}{2}, i \in [1,k]
\end{aligned}
\end{equation}

We incorporate the recurrent visual embedding module into DP-RNN as in Figure~\ref{fig:model}. Given an input image-text pair $(I, T)$, DP-RNN first extracts the word features and the original object features by the text and image encoder. After that, the new object features are extracted by the recurrent visual embedding module in the above process. In the end, the model feeds the word features and the new object features into the Multi-attention Cross Matching Model as described in the above subsection to compute the similarity of pair $(I, T)$. 

\subsection{Pair Early-selection and Training Strategy} \label{sec:learn}
As mentioned in the above subsections, the triplet ranking loss is based on hardest negatives. It means that given a mini-batch that contains $s$ image-text pairs as input, we need to compute the similarity of all $s^{2}$ image-text pairs to decide the hardest negatives. In this process, for each image, there will be $s$ kinds of object orders correspond to $s$ texts. The bi-directional GRU of the recurrent visual embedding module needs to perform a total of $s^{2}$ times of computing for a batch, which sharply increases the model's computing complexity and makes it inapplicable. Therefore, we design an early-selection strategy to filter out insignificant image-text pairs that have no chance to become the hardest negatives.

Given a mini-batch of input texts and images, to filter out the non-corresponding pairs that have no chance to become hardest negatives, we compute an early matching score between each text and each image. For image $I$ and text $T$, the early matching score $S_{em}(I,T)$ is computed as:


\begin{equation}
\begin{aligned}
S_{em}(I,T) = \sum_{i=1}^{k}\sum_{j=1}^{n}P(i,j)
\end{aligned}
\end{equation}
where $P(i,j)$ is the coefficient that $W_{j}$ relates to $O_{i}$ defined in the above subsection. $S_{em}(I,T)$ can be computed by the word features of $T$ and the original object features of $I$. For each text in the mini-batch, we just reserve the top-$d$ non-corresponding images in the same mini-batch with highest early matching score to the text, and compute their similarity by DP-RNN. For each text, we can thus choose the hardest negative image (i.e. the image with highest similarity to the text) from this $d$ images and train DP-RNN by Equation~\ref{equ:loss}. Likewise, for each image, we just choose the hardest negative text (i.e. the text with highest similarity to the image) from the texts that have a computed similarity to the image. This early-selection strategy sharply reduces the computing times of the bi-directional GRU.

We adopt a multi-stage training strategy to train DP-RNN. In the first training epoch, we train the Multi-attention Cross Matching Model, updating the parameters of the text encoder, image encoder (only the fully-connected layer) and attention modules. In the second epoch, we feed the recurrent visual embedding module into the network. We train the parameters of the recurrent visual embedding module with other parameters fixed. From the third epoch to the end, we train the whole network with all the parameters updated.

\section{Experiments}
We perform extensive experiments to evaluate the proposed model. The performance of sentence retrieval (image query) and image retrieval (sentence query) are evaluated by the standard  recall at $K$ (R@K), which is defined as the fraction
of queries for which the correct item belongs to the top-$K$ retrieval items. We first discuss the datasets and model settings used in the experiments. We then compare and analyze the performance of the proposed model with the state-of-the-art image-text matching models.

\subsection{Dateset}
We evaluate our model on the MS-COCO and Flickr30K datasets. Flickr30k \cite{young2014image} consists of 31,783 images collected from the Flickr website. Each image corresponds to five human-annotated sentences. Following the split of \cite{faghri2017vse++,lee2018stacked}, we randomly select 1,000 images for validation and 1,000 images for testing, and use other images to train the model. The original MS-COCO dataset \cite{lin2014microsoft} contains 82,783 training and 40,504 validation images, each image is also annotated with five descriptions. We split the dataset into 82,783 training images, 5,000 validation images and 5,000 test images. Following \cite{faghri2017vse++,lee2018stacked}, we add the 30,504 images that are originally in the validation set but are not used for validating/testing into the training set. Following \cite{wang2019position,hu2019multi}, we report the results by averaging over 5 folds of 1K test images.

\subsection{Implementation Details}

Following \cite{lee2018stacked}, for each image, we extract a $36 \times 2048$ dimensional object features by the Faster R-CNN model in conjunction with ResNet-101 pre-trained by Anderson et al. \cite{anderson2018bottom}. We set the dimensionality of the word embeddings (i.e. $q$) to 300, the dimensionality of image encoder (i.e. $h$) to 1024, and the dimensionality of the bi-directional GRUs for both image and text branches to 1024. The hyper-parameters $\lambda_{1}$, $\lambda_{2}$ are set to 9 and 4, respectively. For MS-COCO, $\beta_{w}$ and $\beta_{o}$ are set to 0.3 and 0 (i.e average the object-text similarity). For Flickr30K, they are set to 0.3 and 0.3. For the recurrent visual embedding module, $d$ is set as 10 for early-selection. The margin of triplet loss $\gamma$ is set to 0.2. We use the Adam optimizer \cite{kingma2014adam} to train the model with the batch size set as 128. For MS-COCO and Flickr30K, the initial learning rate is set to 0.0005 and 0.0002, and is divided by 10 every 10 epochs.

\subsection{Quantitative Results}

\begin{table*}[htbp]
  \small\centering
  \caption{\label{tab:result1} Comparison of the cross-modal retrieval results on MS-COCO. ``Multi-ATT'' represents the Multi-attention Cross Matching Model without recurrent visual embedding. ``DP-RNN (SCAN-based)'' represents the model that directly feeds the proposed recurrent visual embedding module into SCAN \cite{lee2018stacked}. ``DP-RNN (SCAN-based, random order)'' represents the ``DP-RNN (SCAN-based)'' model that receives the object features with random order as input.}
  \begin{threeparttable}
  \resizebox{1.56\columnwidth}{!}{
  \begin{tabular}{|c|c|c|c|c|c|c|}

    \cline{1-7}
    &\multicolumn{3}{|c|}{Sentence Retrieval}&\multicolumn{3}{|c|}{Image Retrieval}\\ 
    Model&R@1&R@5&R@10&R@1&R@5&R@10 \\ \cline{1-7}
    \multicolumn{7}{|c|}{1K Test Images} \\ \cline{1-7}
    DVSA (R-CNN, AlexNet) \cite{karpathy2015deep} &38.4&69.9&80.5&27.2&22.8&74.8\\ 
    HM-LSTM (R-CNN, AlexNet) \cite{niu2017hierarchical}&43.9&-&87.8&36.1&-&86.7\\ 
    Order-embeddings (VGG) \cite{vendrov2015order} &46.7&-&88.9&37.9&-&85.9\\ 
    SM-LSTM (VGG) \cite{huang2017instance} &53.2&83.1&91.5&40.7&75.8&87.4\\
    2WayNet (VGG) \cite{eisenschtat2017linking} &55.8&75.2&-&39.7&63.3&-\\ 
    VSE++ (ResNet) \cite{faghri2017vse++}  &64.6&-&95.7&52.0&-&92.0\\ 
    DPC (ResNet) 
    GXN (ResNet) \cite{gu2018look} &68.5&-&97.9&56.6&-&94.5\\ 
    SCO (ResNet) \cite{huang2018learning} &69.9&92.9&97.5&56.7&87.5&94.0\\
    SCAN (Faster R-CNN, ResNet) \cite{lee2018stacked} &72.7&94.8&98.4&58.8&88.4&94.8\\
    RDAN (Faster R-CNN, ResNet) \cite{hu2019multi}
    &74.6&96.2&98.7&61.6&89.2&94.7\\
    PFAN (Faster R-CNN, ResNet) \cite{wang2019position}
    &\textbf{76.5}&\textbf{96.3}&\textbf{99.0}&61.6&89.6&\textbf{95.2}\\
    \cline{1-7}
    \multicolumn{7}{|l|}{Ours (Faster R-CNN, ResNet):} \\
    DP-RNN (SCAN-based)  &73.8&95.2&98.4&60.5&88.7&94.4\\ 
    DP-RNN (SCAN-based, random order) &72.4&94.7&98.4&59.0&88.4&94.6 \\
    Multi-ATT &73.4&95.4&98.6&60.2&88.8&94.8 \\
    
    DP-RNN &75.3&95.8&98.6&\textbf{62.5}&\textbf{89.7}&95.1\\ \cline{1-7}
    

    \end{tabular}}%
  \end{threeparttable}
\end{table*}%

Table~\ref{tab:result1} shows the quantitative retrieval results of different models on the MS-COCO dataset. From the results, we can see that the proposed DP-RNN remarkably outperforms the baseline SCAN model \cite{lee2018stacked} and has competitive performance compared with other state-of-the-art models. In particular, we separately incorporate different modules we propose to demonstrate the effectiveness. We first feed the proposed recurrent visual embedding module into the original SCAN model (i.e. ``DP-RNN (SCAN-based)'' in Table~\ref{tab:result1}). From the result of ``1K Test Images'', R@1 improves from 72.7 to 73.8 for sentence retrieval, and from 58.8 to 60.5 for image retrieval. When we feed it into our Multi-attention Cross Matching Model, R@1 improves from from 73.4 to 75.3 for sentence retrieval, and from 60.2 to 62.5 for image retrieval. If we shuffle the objects, the performance of the SCAN models with/without the recurrent visual embedding module becomes similar. This demonstrates the effectiveness and validness of recurrent visual embedding on different image-text matching models. 

From Table~\ref{tab:result2}, we can see how different settings of hyper-parameter $d$ influence the performance of DP-RNN. $d$ indicates how many image-text pairs to preserve before we feed them into the recurrent visual embedding modules. Note that the performance drops if $d$ is set to 5. This may be attributed to the wrong discarding of real hard negatives in the pair early-selection process.

Table~\ref{tab:result3} shows the performance of different models on the Flickr30k dataset. Our model outperforms the state-of-the-art approaches for both sentence retrieval
or image retrieval. Overall, on both datasets, our model achieves more improvement for image retrieval. This relates to the essence of the recurrent visual embedding, which changes the object orders based on the text. This attribute fits for the task of ranking images by a reference text query (i.e. image retrieval) because in this process the reference text feature will not be changed when computing its similarity to different images.

\subsection{Qualitative Results}

\begin{figure}[!t]
\centering
\includegraphics[width=0.95\columnwidth]{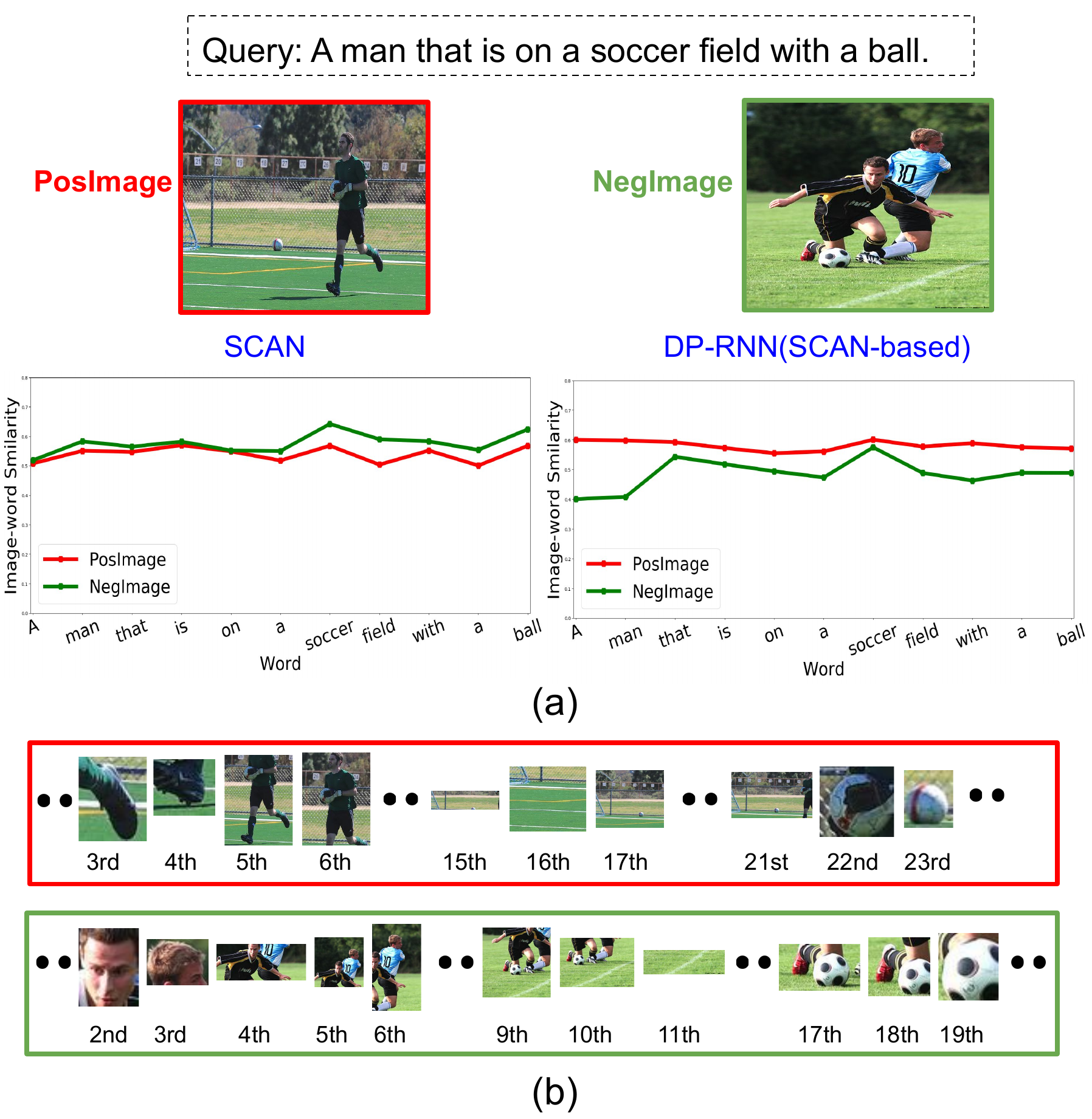}
\caption{On a corresponding image-text pair (PosImage) and a non-corresponding pair (NegImage) sharing the same text, we visualize (a) the image-word similarity $S(I, j)$ predicted by SCAN and DP-RNN (SCAN-based), and (b) the object orders of the two images predicted by DP-RNN (SCAN-based).}
\label{fig:order}
\end{figure}

\begin{figure}[!t]
\centering
\includegraphics[width=0.98\columnwidth]{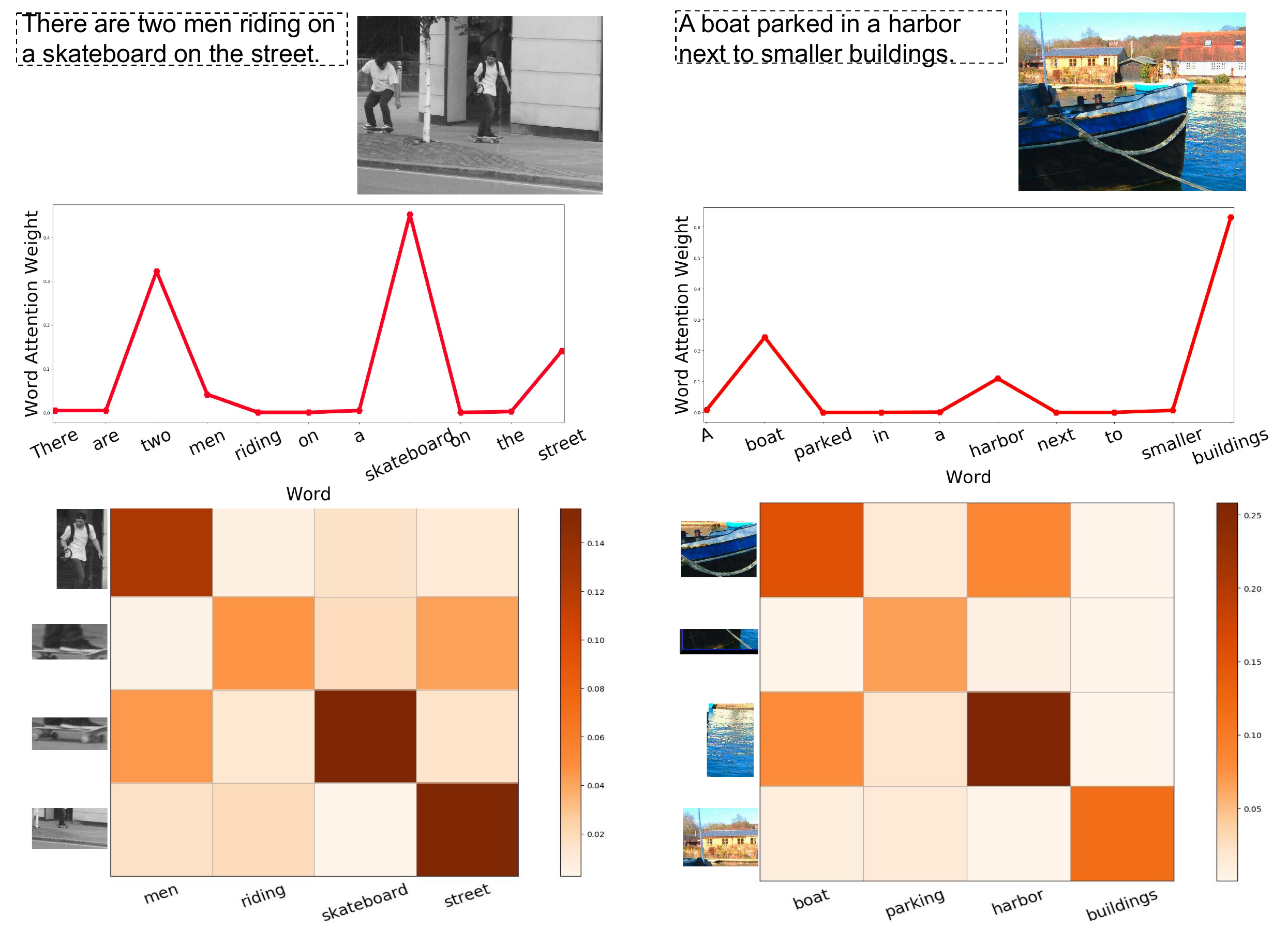}
\caption{Visualization of (a) word attention weight $a_{j}^{w}$ and (b) heat map of word-to-object cross attention weight $\widetilde{\alpha}_{j,i}$ on typical corresponding image-text pairs. $a_{j}^{w}$ and $\widetilde{\alpha}_{j,i}$ are predicted by DP-RNN using the word features and recurrent object features.}
\label{fig:attention}
\end{figure}

\begin{table*}[htbp]
  \small\centering
  \caption{\label{tab:result2} Performance of selecting different $d$ for pair early-selection.}
  \begin{threeparttable}
  \resizebox{1.2\columnwidth}{!}{
  \begin{tabular}{|c|c|c|c|c|c|c|c|}

    \cline{1-8}
    &\multicolumn{3}{|c|}{Sentence Retrieval}&\multicolumn{3}{|c|}{Image Retrieval}&Per Batch \\ 
    Model&R@1&R@5&R@10&R@1&R@5&R@10&Training Time \\ \cline{1-8}
    \multicolumn{8}{|c|}{1K Test Images} \\ \cline{1-8}
    DP-RNN (d=5)  &75.0&95.3&98.3&61.6&89.3&94.6&1.213 \\
    DP-RNN (d=10)  &75.3&95.8&98.6&62.5&89.7&95.1&1.326 \\
    DP-RNN (d=20)  &75.4&95.8&98.7&62.4&89.9&94.9&1.482 \\ \cline{1-8}
    \end{tabular}}%
  \end{threeparttable}
\end{table*}%

\begin{table*}[htbp]
  \small\centering
  \caption{\label{tab:result3} Comparison of the cross-modal retrieval results on Flickr30K.}
  \begin{threeparttable}
  \resizebox{1.5\columnwidth}{!}{
  \begin{tabular}{|c|c|c|c|c|c|c|}

    \cline{1-7}
    &\multicolumn{3}{|c|}{Sentence Retrieval}&\multicolumn{3}{|c|}{Image Retrieval}\\ 
    Model&R@1&R@5&R@10&R@1&R@5&R@10 \\ \cline{1-7}
    HM-LSTM (R-CNN, AlexNet) \cite{niu2017hierarchical}&38.1&-&76.5&27.7&-&68.8\\ 
    SM-LSTM (VGG) \cite{huang2017instance} &42.5&71.9&81.5&30.2&60.4&72.3\\
    2WayNet (VGG) \cite{eisenschtat2017linking} &49.8&67.5&-&36.0&55.6&-\\ 
    DAN (ResNet) \cite{nam2017dual}  &55.0&81.8&89.0&39.4&69.2&79.1\\
    VSE++ (ResNet) \cite{faghri2017vse++}  &52.9&-&87.2&39.6&-&79.5\\ 
    DPC (ResNet) \cite{zheng2017dual}&55.6&81.9&89.5&39.1&69.2&80.9\\ 
    SCO (ResNet) \cite{huang2018learning} &55.5&82.0&89.3&41.1&70.5&80.1\\
    SCAN (Faster R-CNN, ResNet) \cite{lee2018stacked} &67.4&90.3&95.8&48.6&77.7&85.2\\
    RDAN (Faster R-CNN, ResNet) \cite{hu2019multi}
    &68.1&91.0&\textbf{95.9}&54.1&80.9&87.2\\
    PFAN (Faster R-CNN, ResNet) \cite{wang2019position}
    &70.0&\textbf{91.8}&95.0&50.4&78.7&86.1\\
    \cline{1-7}
    DP-RNN &\textbf{70.2}&91.6&95.8&\textbf{55.5}&\textbf{81.3}&\textbf{88.2} \\ \cline{1-7}

    \end{tabular}}%
  \end{threeparttable}
\end{table*}%

\begin{figure*}[!t]
\centering
\includegraphics[width=2.08\columnwidth]{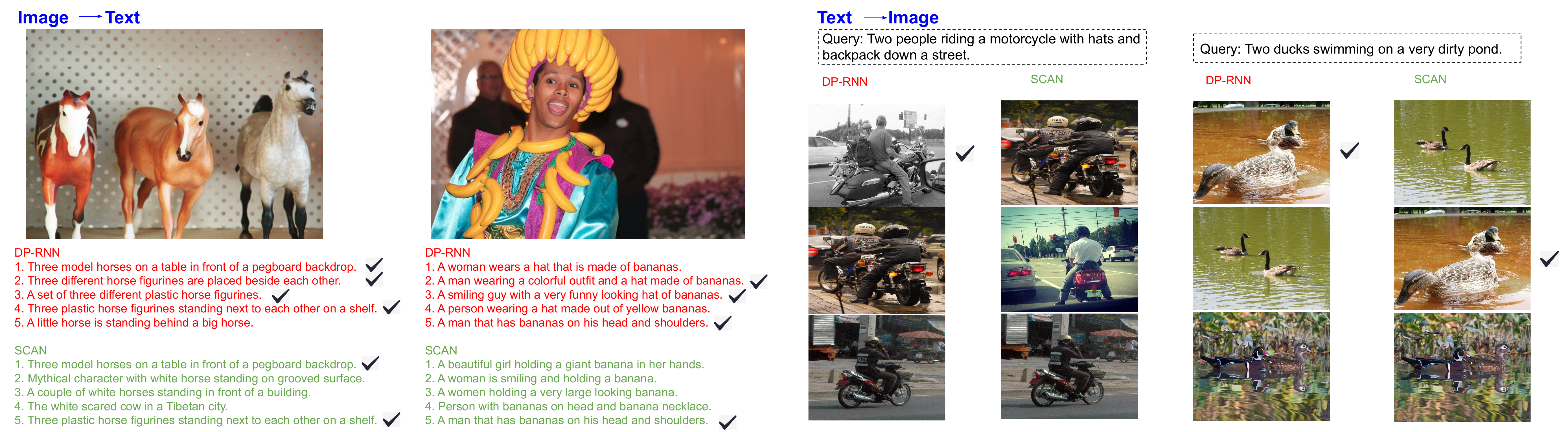}
\caption{Qualitative image retrieval and sentence retrieval comparison between DP-RNN and SCAN on the MS-COCO dataset.}
\label{fig:example}
\end{figure*}

In order to prove that the recurrent visual embedding is effective, we provide examples of the MS-COCO test set to show how recurrent visual embedding has an influence on image retrieval. Given a query ``a man that is on a soccer field with a ball'' with a corresponding image and a non-corresponding image, Figure~\ref{fig:order} illustrates their corresponding object orders predicted by ``DP-RNN (SCAN-based)''. Following the key words ``man'', ``soccer field'' and ``ball'' in the text, the model puts the corresponding objects in the same order as the words for both images. For example, several objects related to people are consecutively put in the first several positions, while objects related to a ball are put in the last several positions. This order helps the model to extract the joint information from semantically related objects. As shown in Figure~\ref{fig:order}, the basic SCAN model predicts close image-word similarity for each word to the two images, falsely predicting higher image-text similarity for the non-corresponding image. The model that feeds the recurrent visual embedding into SCAN predicts a higher image-word similarity for each word to the corresponding image. In particular, it figures out that ``a man'' is not a corresponding phrase to the non-corresponding image after recurrently encoding the objects related to people. 

Figure~\ref{fig:attention} shows the word attention weight $a_{j}^{w}$ and word-to-object cross attention weight $\widetilde{\alpha}_{j,i}$ of typical examples on the MS-COCO test set. Overall, DP-RNN gives key words of each text higher weights, such as ``man'', ``skateboard'', ``street'', ``boat'' and ``buildings''. This strengthens the importance of their corresponding image-word similarity. From the heat map of $\widetilde{\alpha}_{j,i}$, DP-RNN accurately predicts high $\widetilde{\alpha}_{j,i}$ between corresponding word-objects. We could see that the model can even capture the corresponding object of a verb. For example, the region that represents ``the bottom of a boat in the water'' gains highest $\widetilde{\alpha}_{j,i}$ to the word ``parking''.

Figure~\ref{fig:example} shows the qualitative comparison of DP-RNN and SCAN on the MS-COCO test set. For sentence retrieval, given an image query, we show the top-$5$ retrieved sentences predicted by DP-RNN and SCAN. For image retrieval, given a sentence query, we show the top-$3$ retrieved images, ranking from upper to bottom. We tick off the correct retrieval items for each query. It could be seen that DP-RNN performs better in finding key details from visual-textual inputs. Compared with SCAN, it can successfully match the tricky phrases ``hat of bananas'', ``dirty pond'', ``three different horses'' to the corrected images.

\section{Conclusions}
In this paper, we present a Dual Path Recurrent Neural Network for image-text matching. Following the same way as encoding text information, the image objects are reordered adaptively based on their semantic meaning and encoded by RNN. A pair early-selection strategy is proposed to make the recurrent visual embedding tractable. Furthermore, we integrate a Multi-attention Cross Matching Model to compute the image-text similarity from recurrent visual and textual features. We conduct extensive experiments to demonstrate how our model effectively captures the joint information from semantically related objects and match the information to the corresponding words in the text.
\section{Acknowledgment}
This work is partially supported by NSF award \#1704337, and corporate sponsors.

{\small
\bibliographystyle{aaai}
\bibliography{egbib}

\begin{thebibliography}{}

\bibitem[\protect\citeauthoryear{Anderson \bgroup et al\mbox.\egroup
  }{2018}]{anderson2018bottom}
Anderson, P.; He, X.; Buehler, C.; Teney, D.; Johnson, M.; Gould, S.; and
  Zhang, L.
\newblock 2018.
\newblock Bottom-up and top-down attention for image captioning and visual
  question answering.
\newblock In {\em Proceedings of the IEEE Conference on Computer Vision and
  Pattern Recognition},  6077--6086.

\bibitem[\protect\citeauthoryear{Eisenschtat and
  Wolf}{2017}]{eisenschtat2017linking}
Eisenschtat, A., and Wolf, L.
\newblock 2017.
\newblock Linking image and text with 2-way nets.
\newblock In {\em Proceedings of the IEEE conference on computer vision and
  pattern recognition},  4601--4611.

\bibitem[\protect\citeauthoryear{Faghri \bgroup et al\mbox.\egroup
  }{2017}]{faghri2017vse++}
Faghri, F.; Fleet, D.~J.; Kiros, J.~R.; and Fidler, S.
\newblock 2017.
\newblock Vse++: Improved visual-semantic embeddings.
\newblock {\em arXiv preprint arXiv:1707.05612} 2(7):8.

\bibitem[\protect\citeauthoryear{Frome \bgroup et al\mbox.\egroup
  }{2013}]{frome2013devise}
Frome, A.; Corrado, G.~S.; Shlens, J.; Bengio, S.; Dean, J.; Mikolov, T.;
  et~al.
\newblock 2013.
\newblock Devise: A deep visual-semantic embedding model.
\newblock In {\em Advances in neural information processing systems},
  2121--2129.

\bibitem[\protect\citeauthoryear{Gu \bgroup et al\mbox.\egroup
  }{2018}]{gu2018look}
Gu, J.; Cai, J.; Joty, S.; Niu, L.; and Wang, G.
\newblock 2018.
\newblock Look, imagine and match: Improving textual-visual cross-modal
  retrieval with generative models.
\newblock In {\em Proceedings of the IEEE Conference on Computer Vision and
  Pattern Recognition},  7181--7189.

\bibitem[\protect\citeauthoryear{He \bgroup et al\mbox.\egroup
  }{2016}]{he2016deep}
He, K.; Zhang, X.; Ren, S.; and Sun, J.
\newblock 2016.
\newblock Deep residual learning for image recognition.
\newblock In {\em Proceedings of the IEEE conference on computer vision and
  pattern recognition},  770--778.

\bibitem[\protect\citeauthoryear{Hu \bgroup et al\mbox.\egroup
  }{2019}]{hu2019multi}
Hu, Z.; Luo, Y.; Lin, J.; Yan, Y.; and Chen, J.
\newblock 2019.
\newblock Multi-level visual-semantic alignments with relation-wise dual
  attention network for image and text matching.
\newblock In {\em Proceedings of the 28th International Joint Conference on
  Artificial Intelligence},  789--795.
\newblock AAAI Press.

\bibitem[\protect\citeauthoryear{Huang \bgroup et al\mbox.\egroup
  }{2018}]{huang2018learning}
Huang, Y.; Wu, Q.; Song, C.; and Wang, L.
\newblock 2018.
\newblock Learning semantic concepts and order for image and sentence matching.
\newblock In {\em Proceedings of the IEEE Conference on Computer Vision and
  Pattern Recognition},  6163--6171.

\bibitem[\protect\citeauthoryear{Huang, Wang, and
  Wang}{2017}]{huang2017instance}
Huang, Y.; Wang, W.; and Wang, L.
\newblock 2017.
\newblock Instance-aware image and sentence matching with selective multimodal
  lstm.
\newblock In {\em Proceedings of the IEEE Conference on Computer Vision and
  Pattern Recognition},  2310--2318.

\bibitem[\protect\citeauthoryear{Karpathy and Fei-Fei}{2015}]{karpathy2015deep}
Karpathy, A., and Fei-Fei, L.
\newblock 2015.
\newblock Deep visual-semantic alignments for generating image descriptions.
\newblock In {\em Proceedings of the IEEE conference on computer vision and
  pattern recognition},  3128--3137.

\bibitem[\protect\citeauthoryear{Kim \bgroup et al\mbox.\egroup
  }{2016}]{kim2016multimodal}
Kim, J.-H.; Lee, S.-W.; Kwak, D.; Heo, M.-O.; Kim, J.; Ha, J.-W.; and Zhang,
  B.-T.
\newblock 2016.
\newblock Multimodal residual learning for visual qa.
\newblock In {\em Advances in neural information processing systems},
  361--369.

\bibitem[\protect\citeauthoryear{Kingma and Ba}{2014}]{kingma2014adam}
Kingma, D.~P., and Ba, J.
\newblock 2014.
\newblock Adam: A method for stochastic optimization.
\newblock {\em arXiv preprint arXiv:1412.6980}.

\bibitem[\protect\citeauthoryear{Kiros, Salakhutdinov, and
  Zemel}{2014}]{kiros2014unifying}
Kiros, R.; Salakhutdinov, R.; and Zemel, R.~S.
\newblock 2014.
\newblock Unifying visual-semantic embeddings with multimodal neural language
  models.
\newblock {\em arXiv preprint arXiv:1411.2539}.

\bibitem[\protect\citeauthoryear{Lee \bgroup et al\mbox.\egroup
  }{2018}]{lee2018stacked}
Lee, K.-H.; Chen, X.; Hua, G.; Hu, H.; and He, X.
\newblock 2018.
\newblock Stacked cross attention for image-text matching.
\newblock In {\em Proceedings of the European Conference on Computer Vision
  (ECCV)},  201--216.

\bibitem[\protect\citeauthoryear{Li \bgroup et al\mbox.\egroup
  }{2017}]{li2017identity}
Li, S.; Xiao, T.; Li, H.; Yang, W.; and Wang, X.
\newblock 2017.
\newblock Identity-aware textual-visual matching with latent co-attention.
\newblock In {\em Proceedings of the IEEE International Conference on Computer
  Vision},  1890--1899.

\bibitem[\protect\citeauthoryear{Lin \bgroup et al\mbox.\egroup
  }{2014}]{lin2014microsoft}
Lin, T.-Y.; Maire, M.; Belongie, S.; Hays, J.; Perona, P.; Ramanan, D.;
  Doll{\'a}r, P.; and Zitnick, C.~L.
\newblock 2014.
\newblock Microsoft coco: Common objects in context.
\newblock In {\em European conference on computer vision},  740--755.
\newblock Springer.

\bibitem[\protect\citeauthoryear{Lu \bgroup et al\mbox.\egroup
  }{2016}]{lu2016hierarchical}
Lu, J.; Yang, J.; Batra, D.; and Parikh, D.
\newblock 2016.
\newblock Hierarchical question-image co-attention for visual question
  answering.
\newblock In {\em Advances In Neural Information Processing Systems},
  289--297.

\bibitem[\protect\citeauthoryear{Lu \bgroup et al\mbox.\egroup
  }{2017}]{lu2017knowing}
Lu, J.; Xiong, C.; Parikh, D.; and Socher, R.
\newblock 2017.
\newblock Knowing when to look: Adaptive attention via a visual sentinel for
  image captioning.
\newblock In {\em Proceedings of the IEEE Conference on Computer Vision and
  Pattern Recognition (CVPR)}, volume~6.

\bibitem[\protect\citeauthoryear{Mikolov \bgroup et al\mbox.\egroup
  }{2013}]{mikolov2013efficient}
Mikolov, T.; Chen, K.; Corrado, G.; and Dean, J.
\newblock 2013.
\newblock Efficient estimation of word representations in vector space.
\newblock {\em arXiv preprint arXiv:1301.3781}.

\bibitem[\protect\citeauthoryear{Nam, Ha, and Kim}{2017}]{nam2017dual}
Nam, H.; Ha, J.-W.; and Kim, J.
\newblock 2017.
\newblock Dual attention networks for multimodal reasoning and matching.
\newblock In {\em Proceedings of the IEEE Conference on Computer Vision and
  Pattern Recognition},  299--307.

\bibitem[\protect\citeauthoryear{Niu \bgroup et al\mbox.\egroup
  }{2017}]{niu2017hierarchical}
Niu, Z.; Zhou, M.; Wang, L.; Gao, X.; and Hua, G.
\newblock 2017.
\newblock Hierarchical multimodal lstm for dense visual-semantic embedding.
\newblock In {\em Proceedings of the IEEE International Conference on Computer
  Vision},  1881--1889.

\bibitem[\protect\citeauthoryear{Pedersoli \bgroup et al\mbox.\egroup
  }{2017}]{pedersoli2017areas}
Pedersoli, M.; Lucas, T.; Schmid, C.; and Verbeek, J.
\newblock 2017.
\newblock Areas of attention for image captioning.
\newblock In {\em Proceedings of the IEEE International Conference on Computer
  Vision},  1242--1250.

\bibitem[\protect\citeauthoryear{Ren \bgroup et al\mbox.\egroup
  }{2015}]{ren2015faster}
Ren, S.; He, K.; Girshick, R.; and Sun, J.
\newblock 2015.
\newblock Faster r-cnn: Towards real-time object detection with region proposal
  networks.
\newblock In {\em Advances in neural information processing systems},  91--99.

\bibitem[\protect\citeauthoryear{Vendrov \bgroup et al\mbox.\egroup
  }{2015}]{vendrov2015order}
Vendrov, I.; Kiros, R.; Fidler, S.; and Urtasun, R.
\newblock 2015.
\newblock Order-embeddings of images and language.
\newblock {\em arXiv preprint arXiv:1511.06361}.

\bibitem[\protect\citeauthoryear{Wang \bgroup et al\mbox.\egroup
  }{2019}]{wang2019position}
Wang, Y.; Yang, H.; Qian, X.; Ma, L.; Lu, J.; Li, B.; and Fan, X.
\newblock 2019.
\newblock Position focused attention network for image-text matching.
\newblock {\em arXiv preprint arXiv:1907.09748}.

\bibitem[\protect\citeauthoryear{Xu \bgroup et al\mbox.\egroup
  }{2015}]{xu2015show}
Xu, K.; Ba, J.; Kiros, R.; Cho, K.; Courville, A.; Salakhudinov, R.; Zemel, R.;
  and Bengio, Y.
\newblock 2015.
\newblock Show, attend and tell: Neural image caption generation with visual
  attention.
\newblock In {\em International Conference on Machine Learning},  2048--2057.

\bibitem[\protect\citeauthoryear{Yang \bgroup et al\mbox.\egroup
  }{2016}]{yang2016stacked}
Yang, Z.; He, X.; Gao, J.; Deng, L.; and Smola, A.
\newblock 2016.
\newblock Stacked attention networks for image question answering.
\newblock In {\em Proceedings of the IEEE conference on computer vision and
  pattern recognition},  21--29.

\bibitem[\protect\citeauthoryear{You \bgroup et al\mbox.\egroup
  }{2016}]{you2016image}
You, Q.; Jin, H.; Wang, Z.; Fang, C.; and Luo, J.
\newblock 2016.
\newblock Image captioning with semantic attention.
\newblock In {\em Proceedings of the IEEE Conference on Computer Vision and
  Pattern Recognition},  4651--4659.

\bibitem[\protect\citeauthoryear{Young \bgroup et al\mbox.\egroup
  }{2014}]{young2014image}
Young, P.; Lai, A.; Hodosh, M.; and Hockenmaier, J.
\newblock 2014.
\newblock From image descriptions to visual denotations: New similarity metrics
  for semantic inference over event descriptions.
\newblock {\em Transactions of the Association for Computational Linguistics}
  2:67--78.

\bibitem[\protect\citeauthoryear{Yu \bgroup et al\mbox.\egroup
  }{2017}]{yu2017multi}
Yu, D.; Fu, J.; Mei, T.; and Rui, Y.
\newblock 2017.
\newblock Multi-level attention networks for visual question answering.
\newblock In {\em Proceedings of the IEEE Conference on Computer Vision and
  Pattern Recognition},  4709--4717.

\bibitem[\protect\citeauthoryear{Zheng \bgroup et al\mbox.\egroup
  }{2017}]{zheng2017dual}
Zheng, Z.; Zheng, L.; Garrett, M.; Yang, Y.; and Shen, Y.-D.
\newblock 2017.
\newblock Dual-path convolutional image-text embedding with instance loss.
\newblock {\em arXiv preprint arXiv:1711.05535}.

\end{thebibliography}
}

\end{document}